# Chemical-induced Disease Relation Extraction with Dependency Information and Prior Knowledge

Huiwei Zhou[*], Shixian Ning, Yunlong Yang, Zhuang Liu, Chengkun Lang, Yingyu Lin

School of Computer Science and Technology, Dalian University of Technology, Dalian 116024, Liaoning, China

[*] Corresponding author.

Email: zhouhuiwei@dlut.edu.cn, ningshixian@mail.dlut.edu.cn, SDyyl_1949@mail.dlut.edu.cn, zhuangliu1992@mail.dlut.edu.cn, kunkun@mail.dlut.edu.cn, lyydut@sina.com

**Abstract** Chemical-disease relation (CDR) extraction is significantly important to various areas of biomedical research and health care. Nowadays, many large-scale biomedical knowledge bases (KBs) containing triples about entity pairs and their relations have been built. KBs are important resources for biomedical relation extraction. However, previous research pays little attention to prior knowledge. In addition, the dependency tree contains important syntactic and semantic information, which helps to improve relation extraction. So how to effectively use it is also worth studying. In this paper, we propose a novel convolutional attention network (CAN) for CDR extraction. Firstly, we extract the shortest dependency path (SDP) between chemical and disease pairs in a sentence, which includes a sequence of words, dependency directions, and dependency relation tags. Then the convolution operations are performed on the SDP to produce deep semantic dependency features. After that, an attention mechanism is employed to learn the importance/weight of each semantic dependency vector related to knowledge representations learned from KBs. Finally, in order to combine dependency information and prior knowledge, the concatenation of weighted semantic dependency representations and knowledge representations is fed to the softmax layer for classification. Experiments on the BioCreative V CDR dataset show that our method achieves comparable performance with the state-of-the-art systems, and both dependency information and prior knowledge play important roles in CDR extraction task.

## 1 Introduction

The extraction of chemical-disease relation (CDR) provides additional support to precision medicine efforts. It is of essential importance to the clinical disease diagnosis, treatment and drug development [1, 2]. However, manually extracting these relations from biomedical literature into structured knowledge bases, such as Comparative Toxicogenomics Database (CTD) [3], is expensive, time-consuming, and difficult to keep up-to-date. Automatically extracting CDR from the literature is becoming increasingly important for precision medicine as well as drug discovery and basic biomedical research.

To further promote the development of systems for extracting chemical-disease interactions, a challenging task of automatic extraction of CDR from biomedical literature is published on BioCreative V [4]. It consists of two specific subtasks: (i) disease named entity recognition and normalization (DNER) and (ii) chemical-induced diseases (CID) relation extraction. This paper focuses on the CID subtask at both intra- and inter-sentence levels. The intra-sentence level means a given pair of entity mentions is within the same sentence, while the inter-sentence level means a mention pair is in two different sentences.

Since then, much research has been investigated for CDR extraction, such as rule-based methods [5], feature-based methods [6-10]. Lowe et al. [5] propose a simple rule-based system and achieve an *F*-score of 60.75%. Rule-based methods are simple and effective, but difficult to be extended to a new dataset.

Feature-based methods usually achieve better performance than rule-based methods by extracting lexical [7-10], statistic [9, 10], and syntactic features [6-7], [10] etc. Zhou et al. [7] extract structured and flattened dependency features based on the shortest dependency path (SDP) between the chemical and disease entities, which are proved to be effective for CDR extraction. Dependency trees could reflect semantic and syntactic relationships of words in a sentence and achieve better performance than raw word sequences [7]. These elaborately designed features could capture the semantic information and achieve better performance. However, designing and extracting these features is very time-consuming and laborious. And these manually designed features are hard to be migrated into other tasks.

With the recent success of neural networks in natural language processing, different neural networks with complex structures are proposed for learning semantic features from word sequences automatically [11-13]. Zhou et al. [11] propose a hybrid system with a Long Short-term Memory (LSTM) [14] neural network and a tree-kernel-based Support Vector Machine (SVM) to extract semantic and syntactic features respectively, and have achieved 61.31% $F$-score in CDR extraction task on the test dataset with gold standard entity annotations. Li et al. [12] propose a novel convolutional neural network (CNN) [15] model with an attention mechanism for CDR extraction. They perform the convolution operations along the candidate sentences to get the semantic representations and utilize an attention mechanism for the purpose of capturing the important semantic representations. Zheng et al. [13] use an attention mechanism to automatically learn the weight of each hidden representation of the bidirectional LSTM model for classifying drug-drug interactions (DDIs).

The methods mentioned above [11-13] take the whole sentence or the word sequence between the two target entities as input to learn semantic representations. However, for the entity pairs far away from each other, such methods may fail to describe the relation of the two entities, and some irrelevant information may also be considered due to the long distance. To solve these issues, some researchers [16-17] explore dependency trees to capture the crucial semantic dependency information between chemical and disease entities. They utilize the SDPs between chemical and disease pairs as input to extract entity relation, and achieve 61.30% and 65.88% $F$-score respectively on BioCreative V test set with gold standard entity annotations.

In addition to crucial semantic dependency information, knowledge bases (KBs) are also useful for relation extraction. Large-scale KBs usually contain huge amounts of structured triplets as the form of (*head entity*, *relation*, *tail entity*) (also denoted as ($h$, $r$, $t$)). The *relation* indicates the interaction between the head entity and tail entity. Xu et al. [8] introduce many knowledge features derived from several KBs which contain prior knowledge about chemicals and diseases. These features significantly improve the CDR extraction performance from 50.73% to 67.16%. Pons et al. [9] also use rich prior knowledge features, statistical features, and linguistic features, and achieve 70.20% $F$-score on CDR extraction task.

To better model prior knowledge in KBs, some researchers focus on knowledge representation learning, which could learn low-dimensional representations for entities and relations [18-20]. TransE is a typical knowledge representation approach, which represents the relation between the two entities as a translation in a representation space, that is, $\mathbf{h}+\mathbf{r}\approx\mathbf{t}$ when ($h$, $r$, $t$) holds. TransE achieves state-of-the-art prediction performance on 1-to-1 relations, but does not do well in dealing with 1-to-N, N-to-1 and N-to-N relations. Some new methods, such as TransH [19] or TransR [20] have been proposed to model entities or relations in separate entity spaces and relation spaces for solving the problem of 1-to-N, N-to-1 and N-to-N relations. Existing knowledge representation learning methods have been used to extract common entity relations [18-20] based on KBs of general fact such as Freebase [21]. Researchers have built many large-scale KBs in the biomedical area, which are crucial resources for biomedical entity relation extraction. However, knowledge representation learning has not yet been explored in the biomedical entity relation extraction.

This paper aims at applying knowledge representations to CDR extraction, and investigating the effectiveness of the SDP in biomedical text mining. We propose a convolutional attention network (CAN) for CDR extraction. Specifically, we first use Gdep Parser[1] to get the dependency parse trees, and extract the SDP between chemical and disease pairs in a sentence as input. Then, the convolution operations are performed on the SDP to embed deep semantic dependency features. After that, an attention mechanism is used to learn the importance/weight of each semantic dependency vector related to knowledge representations learned from KBs. Finally, for the purpose of combining dependency information and prior knowledge, the weighted semantic dependency representations are concat-enated with knowledge representations and fed into the softmax layer for classification. Experiments on the BioCreative V CDR dataset [4] show that both dependency information and prior knowledge are effective for relation extraction, meanwhile our proposed method achieves comparable results with the state-of-the-art CDR extraction systems. The CAN is only evaluated on CDR task here. It can also be generalized to many other relation extraction tasks such as the common entity relation extraction with Freebase [21], the domain-specific relation extraction of protein-protein interactions (PPIs) with IntAct [22] etc.

---

[1] http://people.ist.usc.edu/~sagae/parser/gdep

The major contributions of this paper are summarized as follows:

(1) We use the convolution operations to capture deep semantic and syntactic information based on the SDP between chemical and disease pairs. Compared with raw word sequences, SDP sequences could provide more concise and effective information for CDR extraction.

(2) Knowledge representations learned from KBs are introduced to CDR extraction. Knowledge representations could serve as an indicator of the entity relation, and significantly improve the performance of CDR extraction.

(3) CAN model could capture the important semantic dependency representations related to the knowledge representations through an attention mechanism, without relying on massive handcrafted features.

## 2 Method

In this section, we introduce our CDR extraction method, which can be divided into six sequential steps.

(1) Generate relation instances according to some heuristic rules and hypernym filtering method at both intra- and inter- sentence levels.
(2) Extract SDP between chemical-disease pairs in relation instances based on the dependency parse trees.
(3) Learn knowledge representations from the knowledge base CTD by TransE model.
(4) Feed the SDPs and the knowledge representations into CAN model for relation extraction at both intra- and inter- sentence levels.
(5) Merge the results of intra- and inter-sentence levels to acquire relations between entities at document level.
(6) Apply some post-processing rules to pick the most likely CDR back, when no relation is found in a document by our model. Then add them to the document level relations to get the final results.

### 2.1 Relation Instance Construction

Relation instances for both training and testing stages should be first constructed. The instances generated from chemical and disease mentions in the same document are pooled into two groups at intra- and inter-sentence levels respectively. The intra-sentence level means the mentions of chemical and disease entities occur in the same sentence, while the inter-sentence level means otherwise. Following Gu et al. [6], [16], some heuristic rules are applied to construct the intra- and inter-sentence levels instances. The details of the heuristic rules are listed as follows.

*2.1.1 Relation instance construction for intra-sentence level*
(1) Only the chemical-disease pairs whose token distance is less than 10 are considered.
(2) If multiple mentions in a sentence refer to the same entity, we only consider the nearest pair of chemical and disease mentions as the instance.
(3) Any mention that occurs in parentheses should be ignored.

*2.1.2 Relation instance construction for inter-sentence level*
(1) Only the entity pairs which are not involved in any intra-sentence level are considered as inter-sentence level instances.
(2) Sentence distance between two mentions in an instance should be less than 3.
(3) If multiple mentions in a document refer to the same entity, the nearest pair of chemical and disease mentions should be kept as the instance.

*2.1.3 Hypernym filtering*
There are hypernym or hyponym relationships between concepts of diseases or chemicals, where a concept was subordinate to another more general concept. However, the goal of CID task is to extract the relationships between the most specific diseases and chemicals. In other words, the relations between hyponym concepts should be considered rather than those between hypernym concepts.

Following Gu et al. [16], we also use the Medical Subject Headings (MeSH) controlled vocabulary [23] to determine the hypernym relationship between entities in a document. Then we remove the hyper-relation instances that involve more general entities than other entities already existing in the candidate instance.

2.2 Shortest dependency path extraction

The shortest dependency path (SDP) between two entities in a dependency parse tree is usually used for entity relation extraction. The semantic dependency among tokens of SDP can offer more concise and effective information for CDR extraction [16, 17]. Taking Sentence 1 as an example, two chemical entities are denoted by wave line and a disease entity is denoted by underline. The disease entity "*seizures*" is associated with the two chemical entities "*cocaine*" and "*benzoylecgonine*".

Sentence 1: <u>Seizures</u> induced by the <u>cocaine</u> metabolite <u>benzoylecgonine</u> in rats.

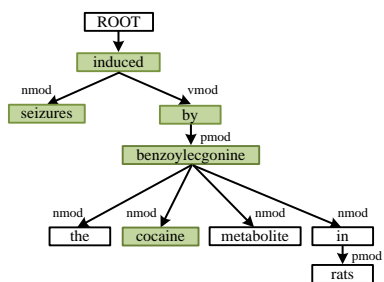

**Fig. 1.** The dependency parse tree of the example sentence. The SDP between the chemical "*cocaine*" and disease "*seizures*" is highlighted in green.

For the dependency parse tree (all words in Sentence 1 are transformed to lowercase) shown in Fig. 1, the SDP between the chemical "*cocaine*" and disease "*seizures*" is highlighted in green. The corresponding directed SDP from the chemical "*cocaine*" to the disease "*seizures*" is shown in Table 1.

**Table 1.** The SDP of an entity pair.

| Chemical | Disease | Shortest dependency path |
|---|---|---|
| *cocaine* | *seizures* | cocaine → nmod → benzoylecgonine → pmod → by → vmod → induced ← nmod ← seizures |

Generally, the SDP could be regarded as a special "sequence". The "tokens" in the path consist of words, dependency directions and dependency relation tags. For example, considering the chemical entity "*cocaine*" and disease entity "*seizures*", the SDP sequence can be represented as {'cocaine', '→', 'nmod', …, 'nmod', '←', 'seizures'}.

Thus, for intra-sentence level instances, we simply extract the SDP from the chemical entity to the disease entity. For inter-sentence level instances, we connect the root node of the dependency tree of the two sentences with an artificially introduced root node. Then we extract the SDP from the chemical entity to the disease entity as inter-sentence level instances.

2.3 Knowledge representation learning

We learn knowledge representations from an expert CDR KB, Comparative Toxicogenomics Database (CTD, update 2017) [3]. Since there are multiple variants of chemical and disease entities, the Medical Subject Headings concept identifiers (MeSH ID) [23] are used to identify chemicals and diseases instead of using the entity mentions.

*2.3.1 Triple extraction*

First, we extract all the chemical-disease pairs both in the CDR dataset (all positives and negatives in training, development and test dataset) and CTD. Then, the relations of the chemical-disease pairs are extracted from CTD to get CID triples (note that the relations in the triples are built on CTD itself, rather than the label of CDR dataset). There are three kinds of relations in CTD: "*inferred-association*", "*therapeutic*", "*marker/mechanism*". Taking the entity pair in Sentence 1 to explain, the relationship of the chemical-disease pair *cocaine* (MeSH ID: D003042), *seizures* (MeSH ID: D012640) in CTD is "*marker/mechanism*". Then it can form a triplet $(e_c, r, e_d)$ (D003042, *marker/mechanism*, D012640), where $e_c$ indicates a *chemical* entity, $r$ is the relationship between the two entities, and $e_d$ indicates a *disease* entity.

For the chemical-disease pairs whose relationships cannot be found in CTD, we simply introduce a special relationship "*null*" to make up these uncovered entity pairs. For example, the training instance pair (D013390, D013746) is not found in CTD. We complete this triple as (D013390, *null*, D013746). Finally, we acquire four kinds of relations: "*inferred-association*", "*therapeutic*", "*marker/mechanism*" and "*null*" of 14261 different chemicals, 5862 diseases and around 1.8 million triplets.

*2.3.2 Learning knowledge representation by TransE*

In this paper, with simplicity and good performance in mind, TransE is selected to learn knowledge representations. All the generated triples are regarded as correct triplets to learn simultaneously chemical representations $\mathbf{e}_c$, disease representations $\mathbf{e}_d$ and relation representations $\mathbf{r}$ in the vector space $\mathbb{R}^k$ by TransE.

In order to combine knowledge representations and word representations into the same vector space. We use the MeSH ID to represent the entity and employ Word2Vec [24] to pre-train entity representations and word representations together on the PubMed articles provided by Wei et al. [25]. All the chemical and disease entities in the articles are recognized and tagged automatically with their corresponding MeSH ID by PubTator [25].

The pre-trained entity representations are then used as the initial entity representations for TransE training. The loss function of TransE is defined as:

$$L = \sum_{(e_c, r, e_d) \in S} \sum_{(e'_c, r, e'_d) \in S'} \max(0, \gamma - \|\mathbf{e}_c + \mathbf{r} - \mathbf{e}_d\| + \|\mathbf{e}'_c + \mathbf{r} - \mathbf{e}'_d\|) \quad (1)$$

where $\gamma$ is the margin, $S$ is the set of correct triplets and $S'$ is the set of incorrect triplets. Since CTD only contains correct triples $(e_c, r, e_d) \in S$, we follow Wang et al. [19] to replace the chemical or disease entity to build the negative triples $(e'_c, r, e_d)$ or $(e_c, r, e'_d)$.

The relation representations learned by TransE are then introduced to our CAN model to find the important evidences from SDP sequences.

2.4 Relation extraction

Fig. 2 illustrates the framework of our CAN model. For a relation instance, the model takes the relation representation learned by TransE and the SDP sequence from chemical entity to disease entity as input. CAN could find out the potential semantic dependency information in the SDP and integrate prior knowledge well. It primarily consists of four layers: a representation layer, a convolutional layer, a knowledge-based attention layer and a softmax layer.

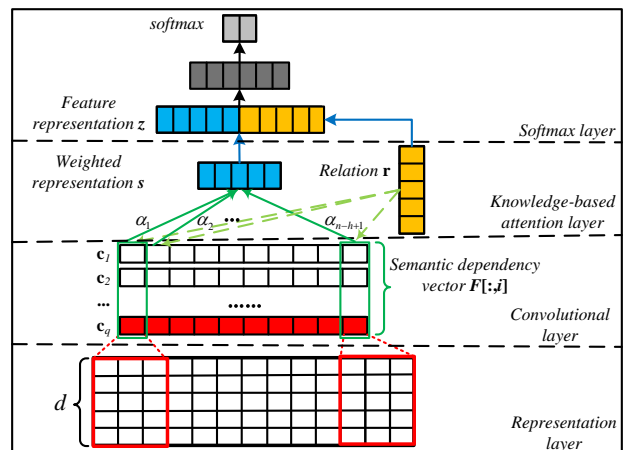

**Fig. 2.** The convolutional attention network for CDR extraction.

In the representation layer, a SDP sequence is represented as a matrix representation with $d$-dimensional vector $x_i \in \mathbb{R}^d$ that represents the $i$-th token in the sequence.

In the convolutional layer, we apply the convolution operations over the input matrix to capture deep semantic dependency features.

In the knowledge-based attention layer, an attention mechanism is used to learn the importance/weight of each semantic dependency vector with regard to the relation representation learned by TransE.

In the softmax layer, the weighted sum of the semantic dependency vectors and the relation representation of the entity pair are concatenated as the final feature representations, which are then fed into a 2-layer perceptron to predict the entity relation by a softmax function.

*2.4.1 Representation layer*

Given a SDP sequence $w = \{w_1, w_2, ..., w_n\}$ of a candidate instance, we map each token into its vector representation to obtain a matrix representation $x = \{x_1, x_2, ..., x_n\} \in \mathbb{R}^{d \times n}$, where $x_i \in \mathbb{R}^d$ is a $d$-dimensional vector. Similarly, the relation $r$ of the entity pair is also mapped into its vector representation $\mathbf{r} \in \mathbb{R}^k$ which is learned by TransE. Then we use the convolutional layer to extract the semantic dependency features.

### 2.4.2 Convolution layer

In this layer, a set of filters of different sizes are used to slide over the candidate matrix representation $x = \{x_1, x_2, ..., x_n\}$ and compute the dot product to obtain multiple feature maps. Let $x_{i:i+h-1}$ refer to the concatenation of the context window of $h$ tokens from $x_i$ to $x_{i+h-1}$. Given a filter $w \in \mathbb{R}^{h \times d}$ of size $h$, the convolution operations can be expressed as $c_i = f(w \bullet x_{i:i+h-1} + b)$. Here $b$ is a bias term and $f$ is a non-linear function such as the rectified linear units (relu) [26]. Fig. 2 shows an example of filter size $h = 3$. Each filter is used for each possible window of tokens in the sequence $\{x_{1:h}, x_{2:h+1}, ..., x_{n-h+1:n}\}$ to produce a feature map: $\mathbf{c} = [c_1, c_2, ..., c_{n-h+1}]$ with $\mathbf{c} \in \mathbb{R}^{n-h+1}$. In this paper, we use $q$ filters to obtain multiple feature maps $F = [\mathbf{c}_1, \mathbf{c}_2, ..., \mathbf{c}_q]^T \in \mathbb{R}^{q \times (n-h+1)}$. The convolution operations could capture the deep semantic features among the SDP tokens. The column $i$ in $F$ is defined as a semantic dependency vector $F[:,i]$ as shown in the green box in Fig. 2.

### 2.4.3 Knowledge-based attention layer

In CDR extraction, the indicative words or phrases are the most meaningful information that the model should pay attention to. For example, some trigger words such as "induced", "caused" and some trigger phrases such as "be induced by", "caused by" should have the bigger weight if a sentence expresses the chemical-induced disease relations. Following this intuition, the attention mechanism is employed to learn the importance/weight of each semantic dependency vector with respect to the relation representations learned from KBs.

For each semantic dependency vector $F[:,i]$ gotten by the convolution operations on a candidate instance, we use a feed forward network to compute its semantic relatedness with the relation representation $\mathbf{r} \in \mathbb{R}^k$ of the candidate instance as:

$$g_i = tanh(W_a(F[:,i] \oplus \mathbf{r}) + b_a) \quad (2)$$

where $\oplus$ denotes the concatenation operation, $W_a \in \mathbb{R}^{1 \times (q+k)}$ is the attention matrix and $b_a \in \mathbb{R}^{1 \times 1}$ is the bias.

After obtaining $\{g_1, g_2, ..., g_{n-h+1}\}$, the attention weight of each semantic dependency vector can be defined with a softmax function as follows:

$$\alpha_i = \frac{\exp(g_i)}{\sum_{j=1}^{n-h+1} \exp(g_j)} \quad (3)$$

Then the weighted semantic dependency representation $s$ is defined as follows:

$$s = \sum_{i=1}^{n-h+1} \alpha_i F[:,i] \quad (4)$$

Inspired by Tay et al. [27], to make full use of prior knowledge, the weighted semantic dependency representation is concatenated with the relation representation as the final feature representation $z \in \mathbb{R}^{q+k}$:

$$z = s \oplus \mathbf{r} \quad (5)$$

This simple operation can improve the performance of relation extraction and we will show our results in section 3. An intuitive explanation for the concatenation is that the weighted semantic dependency representation contains useful dependency information driven from SDP. Meanwhile, the relation representation contains prior knowledge driven from CTD. The combination of dependency information and prior knowledge could improve the system performance.

### 2.4.4 Softmax layer

The final feature representation $z \in \mathbb{R}^{q+k}$ is fed to a 2-layer perceptron. We take the non-linear transformation of relu as the activation function. The transformations can be written as follows:

$$h_1 = relu(Wz + b) \quad (6)$$

$$h_2 = relu(W_{h1}h_1 + b_{h1}) \quad (7)$$

where $W \in \mathbb{R}^{n_{h1} \times (d+k)}, W_{h1} \in \mathbb{R}^{n_{h2} \times n_{h1}}, b \in \mathbb{R}^{n_{h1}}, b_{h1} \in \mathbb{R}^{n_{h2}}$ are the parameters that need to be trained.

During the training step, we adopt dropout operation to prevent the over-fitting problem of the 2-layer perceptron, by randomly setting the elements in $h_1$ and $h_2$ to zero by a proportion $p$. And the hidden representations are obtained accordingly:

$$h_1 = dropout(h_1 \odot m_1) \quad (8)$$

$$h_2 = dropout(h_2 \odot m_2) \quad (9)$$

where $\odot$ is an element-wise multiplication and $m_1, m_2$ are the mask embeddings whose elements follow the Bernoulli distribution with the proportion $p$.

Finally, the hidden representation $h_2$ is fed to a softmax function to compute the confidence of CDR:

$$o = softmax(\mathbf{W}_o h_2 + b_o) \quad (10)$$

where $o \in \mathbb{R}^{n_o}$ is the output, $\mathbf{W}_o \in \mathbb{R}^{n_o \times n_{h2}}$ is the weight matrix and $b_o \in \mathbb{R}^{n_o}$ is the bias.

2.5 Relation Merging

After relation extraction, the results of the intra- and inter-sentence level are merged as the final document level results. However, since there may be multi-instances of the same entity pair in a document, it is possible that they are predicted inconsistently [6]. If at least one instance is predicted as positive by our model for the same entity pair, we would believe this entity pair has the true CID relation.

2.6 Post-processing

To further improve the performance, we use some post-processing rules [11] to help extract relations when no CDR is found in a document by the CAN model. The rules are listed as follows:

(1) All the chemicals in the title are associated with all the diseases in the entire document.
(2) When there is no chemical in the title, the most frequently mentioned chemical in the document is associated with all diseases in the entire document.

## 3 Experiments and Results

### 3.1 Experimental Setup

**Dataset.** The CDR dataset released by BioCreative V task [4] is used to evaluate our method, which contains a total of 1500 PubMed articles: 500 each for the training, development and test set. It is an annotated text corpus that consists of human annotations for all chemicals, diseases and CID relations at the document level. Table 2 describes the details of the dataset. The chemical and disease mention columns are the number of total mentions. The ID and CID columns are the total number of different MeSH ID or CID relations.

Table 2. Statistics of the CDR dataset.

| Task Data | Articles | Chemical | | Disease | | CID[+] |
|---|---|---|---|---|---|---|
| | | Men[*] | ID | Men[*] | ID | |
| Training | 500 | 5203 | 1467 | 4182 | 1965 | 1038 |
| Development | 500 | 5347 | 1507 | 4244 | 1865 | 1012 |
| Test | 500 | 5385 | 1435 | 4424 | 1988 | 1066 |

Men[*]: Mention, ID: MeSH ID, CID[+]: CID relations

**Evaluation metrics.** In our experiments, sentences in the corpus are preprocessed with Gdep Parser[2] to get the dependency trees. Following Zhou et al. [11], the original training set and development set are merged as the training set. We randomly select 20% of the training set as a validation set to tune the hyper-parameters and test our model on the test set with the golden standard entities. The evaluation is reported by the official evaluation toolkit[3], which adopts the commonly used metrics of the Precision (*P*), Recall (*R*) and *F*-score (*F*).

**Embedding initialization.** Entity representations and word representations are pre-trained on the PubMed articles[4] provided by Wei [25] with Word2Vec [24]. All articles consist of 27 million documents, 3.4 billion tokens, and 4.2 million distinct words. The dimensions of word, entity and relation representations are all 100. Note that for those words and entities that do not occur in the pre-training corpus (PubMed articles [25]), we take a random embedding with the uniform distribution in $[-0.25, 0.25]$ to initialize them. The dependency relation tags and directions representations are randomly initialized and adapted during training.

**Model hyperparameter settings.** Except that the epoch is set to 500, the other parameters of TransE are consistent with the code released by Lin et al [20]. CAN is trained by Adam technique [28] and L2-norm regularization with parameter 0.0001. The mini-batch size is 32. 100 filters with window size $h = 1, 2, 3, 4$ respectively are used in the convolutional layer. The dimensions of 2-layer perceptron in the softmax layer are 100 and 50 with the same dropout proportion $p=0.5$ respectively. Our CAN model is implemented with an open-source deep learning library Keras [29]. You could find the source code at https://github.com/Xls1994/CDRextraction.

---
2  http://people.ist.usc.edu/~sagae/parser/gdep
3  http://www.biocreative.org/tasks/biocreative-v/track-3-cdr/
4  ftp://ftp.ncbi.nlm.nih.gov/pub/lu/PubTator/

3.2 Method comparison

In the experiments, we first compare the proposed CAN with the following four baseline methods. We also explain how SDP sequences achieve better performance than raw word sequences between the two candidate entities.

(1) **TransE:** This is a naive method of relation extraction with KBs. For a pair of entities in a document, we calculate the cosine similarities between $\mathbf{e}_d - \mathbf{e}_c$ and four kinds of relation representations $\mathbf{r}$ respectively. Then we rank the relations according to the four corresponding cosine similarities. Only the entity pairs with top-1 relationship "*marker/mechanism*" are considered to have the true CID relations. **TransE** merely uses the entity and relation representations and do not need context information.

(2) **CNN:** This method applies convolution and max pooling operation along the SDP sequences. In practice, 100 filters with different filter sizes $h=\{1,2,3,4\}$ respectively are used to obtain a set of different feature maps. The **CNN** model is similar to our **CAN** model except that **CNN** uses max pooling rather than uses knowledge-based attention mechanism to get semantic dependency representations for classification.

(3) **LSTM:** This method encodes SDP sequences with long-short term memory networks (LSTM). The dimension of hidden representation is 100 and the final hidden representation of the LSTM is used for classification.

(4) **LSTM-KA:** This method applies LSTM with knowledge-based attention (KA) mechanism, which is most similar to our CAN method among the four baseline methods. Their main difference is that **LSTM-KA** use the hidden representations of each time step of LSTM to calculate the attention score instead of using the semantic dependency vector $F[:,i]$ gotten by the convolution operations.

Table 3 and Table 4 show the comparison results at intra- and inter-sentence level respectively. Seen from the two tables, we find that:

(1) All methods with SDP sequences have higher *F*-score than that with raw word sequences. This suggests that SDP could capture the most direct syntactic and semantic information of the two entities and provide strong hints for CDR extraction.

(2) **TransE** with pure knowledge information performs poorly due to the lack of effective context information. This shows that context information is indispensable to CDR extraction.

(3) Among the four baseline methods, the best one is **LSTM-KA**, which is similar to **CAN** except it calculates semantic dependency features by LSTM instead of the convolution operations. Compared with LSTM, which is suitable to learn long terms dependencies, the convolution operations are suitable to capture the local features. In most cases, relations are predominantly reflected in local features rather than in global features.

(4) Furthermore, compared with **CAN**, the performance of **CNN** has dropped significantly without the help of knowledge representations. Knowledge representations could provide effective prior knowledge of the chemical-disease pairs and significantly improve the performance.

Overall, dependency information and prior knowledge are both beneficial to CDR extraction.

**Table 3.** Comparison with baseline methods on the test dataset at intra-sentence level.

| Method | Word Sequence | | | SDP Sequence | | |
|---|---|---|---|---|---|---|
| | *P* (%) | *R* (%) | *F* (%) | *P* (%) | *R* (%) | *F* (%) |
| **TransE** | 43.83 | 32.00 | 37.00 | - | - | - |
| **CNN** | 50.60 | 55.16 | 52.79 | 56.50 | 52.15 | 54.24 |
| **LSTM** | 53.13 | 49.25 | 51.12 | 54.50 | 52.91 | 53.69 |
| **LSTM-KA** | 64.55 | 61.16 | 62.81 | 62.08 | 65.10 | 63.55 |
| **CAN** | 62.96 | 63.79 | 63.37 | 65.90 | 62.20 | 64.00 |

**Table 4.** Comparison with baseline methods on the test dataset at inter-sentence level.

| Method | Word Sequence | | | SDP Sequence | | |
|---|---|---|---|---|---|---|
| | P (%) | R (%) | F (%) | P (%) | R (%) | F (%) |
| **TransE** | 19.86 | 13.79 | 16.28 | - | - | - |
| **CNN** | 32.31 | 4.97 | 8.61 | 42.65 | 8.16 | 13.70 |
| **LSTM** | 26.29 | 4.78 | 8.10 | 30.03 | 8.26 | 12.95 |
| **LSTM-KA** | 44.73 | 13.13 | 20.30 | 50.35 | 13.23 | 20.95 |
| **CAN** | 55.56 | 12.83 | 20.85 | 49.65 | 13.60 | 21.35 |

### 3.3 Effects of the post-processing

The results of the intra- and inter-sentence level are merged as the document level results. To further improve the extraction performance, we apply some post-processing rules to the document level result of **CAN**, and the results after relation merging and post-processing is shown in Table 5. From the table, we can find that the recall increases significantly from 75.80% to 80.48% while the precision decreases slightly compared with the result of relation merging. The post-processing could help the **CAN** to pick the most likely CDR back when no CDR is found in a document. As a supplement to the system, the post-processing has a very strong effect on the CDR extraction.

Noting that, the recall of relation merging is the sum of intra- and inter- sentence levels. The reason is perhaps that the intra- and inter- sentence instances are totally irrelevant due to our heuristic rules in section 2.1. Hence, there are not the same CID relations between intra- and inter- sentence level results.

**Table 5.** Result of the post-processing.

| Method | P (%) | R (%) | F (%) |
|---|---|---|---|
| Inter-sentence level | 49.65 | 13.60 | 21.35 |
| Intra-sentence level | 65.90 | 62.20 | 64.00 |
| Relation merging | 62.41 | 75.80 | 68.45 |
| Post-processing | 60.51 | 80.48 | 69.08 |

## 4 Discussion

### 4.1 Effects of knowledge representation learning

The effects of knowledge representation learning are shown in Table 6. **Random** relation representations are randomly initialized with the uniform distribution in $[-0.25, 0.25]$ and then fine-tuned during the training phase.

It can be seen from Table 6 that using relation representations learned by **TransE** outperforms **Random** at both intra- and inter- sentence levels, regardless of **LSTM-KA** or **CAN**. Knowledge representations learned from KBs could improve the intra-sentence level F-score from 62.94% to 64.00%, and the inter-sentence level F-score from 18.65% to 21.35%, in the case of **CAN**. This indicates that **TransE** could capture deep knowledge representations and provide more exact information than **Random**.

### 4.2 Effects of knowledge-based attention mechanism

To explore the effects of our knowledge-based attention (KA) mechanism, **CAN** is compared with the following variants which all use the same SDP sequences as **CAN**.

(1) **attCNN:** This method learns the importance of each semantic dependency vector obtained by the convolution operations without concatenating the relation representation by: $g_i = tanh(W_a F[:,i] + b_a)$. And the weighted semantic dependency representation $s = \sum_{i=1}^{n-h+1} \alpha_i F[:,i]$ is directly fed to the softmax layer without concatenating the relation representation.

(2) **attCNN (Diff):** This method adopts $r_e = v_d - v_c$ as the relation representation, instead of using the relation representation **r** learned by TransE, to weight each semantic dependency vector by attention operation. Here, the chemical representation $v_c$ and disease representation $v_d$ are both learned by Word2Vec. The attention weight is calculated by $g_i = tanh(W_a (F[:,i] \oplus r_e) + b_a)$. And the weighted semantic dependency representation $s = \sum_{i=1}^{n-h+1} \alpha_i F[:,i]$ is directly fed to the softmax layer without concatenating the relation representation.

(3) **attCNN (RR):** This method uses both the relation representation and the semantic dependency vector to get the attention scores $g_i = tanh(W_a (F[:,i] \oplus \mathbf{r}) + b_a)$ as **CAN** does. And the weighted semantic dependency representation $s = \sum_{i=1}^{n-h+1} \alpha_i F[:,i]$ is directly fed to the softmax layer without concatenating the relation representation.

Table 7 shows the performance of different attention mechanisms at both intra- and inter- sentence level. From Table 7, we can observe that:

(1) **attCNN (Diff)** gets better results than **attCNN**, which illustrates the attention mechanism could capture some useful information between the semantic dependency representations and the difference vector $r_e$. This information makes it easier to predict the relations between two entities.

(2) With the help of relation representations learned by TransE, **attCNN (RR)** significantly outperforms **attCNN (Diff)**. TransE enables the establishment of complex semantic relationships between entities and relations, resulting in better relation representations.

(3) **CAN** gets a better result than **attCNN (RR)**. According to Tay et al. [27], it may be hard for attention mechanism to model the knowledge representation and the semantic dependency representation at the same time. **CAN** uses a simple concatenation operation to enhance the stability and reliability of the model. Compared with **attCNN (RR)**, **CAN** could better integrate dependency information and prior knowledge together.

**Table 6.** Effects of knowledge representation learning.

| Relation Representation Initialization | Method | Intra-sentence level | | | Inter-sentence level | | | Relation merging | | |
|---|---|---|---|---|---|---|---|---|---|---|
| | | $P$ (%) | $R$ (%) | $F$ (%) | $P$ (%) | $R$ (%) | $F$ (%) | $P$ (%) | $R$ (%) | $F$ (%) |
| **Random** | LSTM-KA | 61.42 | 64.54 | 62.95 | 52.63 | 12.19 | 19.80 | 59.96 | 76.73 | 67.32 |
| | CAN | 63.51 | 62.38 | 62.94 | 48.62 | 11.53 | 18.65 | 60.90 | 73.93 | 66.78 |
| **TransE** | LSTM-KA | 62.08 | 65.10 | 63.55 | 50.35 | 13.23 | 20.95 | 60.03 | 78.33 | 67.97 |
| | CAN | 65.90 | 62.20 | 64.00 | 49.65 | 13.60 | 21.35 | 62.41 | 75.80 | 68.45 |

**Table 7.** Effects of knowledge-based attention mechanism.

| Method | Intra-sentence level | | | Inter-sentence level | | | Relation merging | | |
|---|---|---|---|---|---|---|---|---|---|
| | $P$ (%) | $R$ (%) | $F$ (%) | $P$ (%) | $R$ (%) | $F$ (%) | $P$ (%) | $R$ (%) | $F$ (%) |
| **attCNN** | 58.97 | 47.47 | 52.60 | 32.17 | 7.79 | 12.54 | 53.06 | 55.25 | 54.14 |
| **attCNN (Diff)** | 51.93 | 56.85 | 54.28 | 33.23 | 10.04 | 15.42 | 48.24 | 66.89 | 56.05 |
| **attCNN (RR)** | 59.91 | 65.20 | 62.44 | 49.81 | 12.48 | 19.95 | 58.30 | 77.68 | 66.61 |
| **CAN** | 65.90 | 62.20 | 64.00 | 49.65 | 13.60 | 21.35 | 62.41 | 75.80 | 68.45 |

### 4.3 Comparison with related works

We compare our results with some relevant systems of the BioCreative V CDR Task in Table 8. In order to make a fair comparison with every system and eliminate the influence of the accumulated errors introduced by different named entity recognition tools, all the systems are reported on the test dataset with the golden standard entity annotations. We mainly divide these relevant systems into three groups: Rule-based methods and Machine Learning-based methods with or without additional resources, namely ML with KBs and ML without KBs.

**Table 8.** Comparison with related works.

| Method | System | $P$ (%) | $R$ (%) | $F$ (%) |
|---|---|---|---|---|
| Rule-based | Lowe et al. [5] | 59.29 | 62.29 | 60.75 |
| ML without KBs | Gu et al. [6] | 62.00 | 55.10 | 58.30 |
| | Gu et al [16] | 55.70 | 68.10 | 61.30 |
| | Zhou et al. [11] | 55.56 | 68.39 | 61.31 |
| | Le et al. [17] | 58.02 | 76.20 | 65.88 |
| ML with KBs | Peng et al. [10] | 68.15 | 66.04 | 67.08 |
| | Xu et al. [8] | 65.80 | 68.57 | 67.16 |
| | Li et al. [12] | 59.97 | 81.49 | 69.09 |
| | Pons et al. [9] | 73.10 | 67.60 | 70.20 |
| | **Ours** | 60.51 | 80.48 | 69.08 |

From Table 8, we can see that the Rule-based methods, i.e. Lowe et al. [5] achieve a comparable performance with most ML without KBs methods. However, these handcrafted rules are difficult to apply to a new dataset.

ML without KBs methods perform better than Rule-based methods relatively, among which Le et al. [17] exploit the SDP between disease and chemical entities with CNN, and finally achieve the highest *F*-score of 65.88%. The difference of SDP sequences and raw word sequences is considerable.

ML with KBs methods significantly outperform the Rule-based methods and ML without KBs methods. Among them, Peng et al. [10] extract one-hot knowledge features based on CTD and MeSH databases, and achieve an *F*-score of 67.08%. Xu et al. [8] use four freely available large-scale prior knowledge bases to extract the prior knowledge features, which contribute 16.43% *F*-score to CDR extraction performance according to their reports. Pons et al. [9] use a commercial system named Euretos Knowledge Platform, which contains entities and relations from structured databases, such as UniProt, CTD and UMLS. Besides prior knowledge features extracted from the Euretos Knowledge Platform, they also extract statistical and linguistic features from the document. Finally they achieve the best performance with an *F*-score of 70.20%. Compared with these ML with KBs methods [8-10], our method does not require extensive manual feature engineering and would be more universal and easier to apply.

Particularly, Li et al. [12] propose a CNN model with attention mechanism, which is the most relevant to our system. Though it achieves a similar result compared with Li et al. [12], our system has some difference from their system. 1) Li et al. [12] take the whole sentence as input, while we use the SDP sequence. 2) Li et al. [12] extract prior knowledge from the four domain knowledge bases CTD, MeSH, MEDication Indication Resource and Side Effect Resource [8]. According to Xu's [8] report, more prior knowledge could achieve better performance. However, we only use CTD to train knowledge representations and achieve a comparable result. 3) Li et al. [12] utilize the one-hot knowledge features. We use knowledge representations learned by TransE to help selecting the most important evidence in the semantic dependency features.

In summary, our method could grasp deep semantic dependency representations with respect to the knowledge representations, and do better in integrating the dependency information and prior knowledge than Li et al. [12].

4.4 Statistical significance test of different methods

To see whether our method yields a significant difference, *t*-test statistics are performed by 10-fold cross validation on the training and development datasets. The average *F*-score improvement of method 1 compared to method 2 and the *P*-values is given in Table 9. **CAN with Random** uses the random relation representations while **CAN with TransE** uses the relation representations learned by TransE.

**Table 9** Statistical significance of performance over 10-fold cross validation.

| Method 1 | Method 2 | Average *F*-score improvement (%) | *P*-values |
|---|---|---|---|
| **CAN with Random** | CNN | 11.62 | 9.66E-07 |
| **CAN with TransE** | CNN | 13.1 | 1.55E-06 |
| **CAN with TransE** | CAN with Random | 1.47 | 0.28E-01 |
| **CAN with TransE** | attCNN (RR) | 1.6 | 0.18E-01 |

From Table 9, we can see *t*-test for **CAN with Random** vs. **CNN** and **CAN with TransE** vs. **CNN** results *P*-value of 9.66E-07 and 1.55E-06 respectively, which shows a significant difference after knowledge representations were introduced. Furthermore, difference between **CAN with TransE** and **CAN with Random** is also significant (*P*-value<0.05), which indicates that the relation representations learned by TransE outperforms random initialization significantly. Compared with **attCNN (RR)**, **CAN with TransE** is also significant (*P*-value<0.05), which indicates that concatenating the relation representations to the weighted semantic dependency representations could improve the performance.

4.5 Error analysis

We perform an error analysis of the final results to detect the origins of false positives (FPs) and false negatives (FNs).

For FPs, two main error types are listed as follows:

(1) Incorrect classification: In spite of the detailed semantic representations, 483 FPs errors come from the in-

correct classification made by our CAN model. Among the 483 FPs, 343 FPs come from the intra-sentence level and 140 FPs come from the inter-sentence level.
(2) Post-processing error: The post-processing rules bring 77 false CDR, with a proportion of 13.75%. That is, some of the negative instances are misclassified into positive instances.

For FNs, two main error types are listed as follows:
(1) Missed classified relation: 47 inter-sentence level instances are removed by the heuristic rules mentioned in the section **2.1.2 Relation instance construction for inter-sentence level**, which are not classified by our CAN model at all because the sentence distance between these chemical and disease entities are more than 3.
(2) Incorrect classification: Our model misclassifies 161 positive cases (67 intra-sentence level positive cases and 94 inter-sentence level positive cases) as negatives due to complex syntactic and latent semantic information of entity pairs.

## 5 Conclusion

We propose a novel convolutional attention network (CAN) for CDR extraction, which achieves 69.08% $F$-score on the BioCreative V CDR task. Our model performs the convolution operations over the shortest dependency path (SDP) between chemical and disease pairs to produce deep semantic dependency features. And then, an attention mechanism is employed to capture the weighted semantic dependency representations related to knowledge representations learned from KBs. Experimental results verify that the proposed approach is comparable to the state-of-the-art feature-based methods. Our results further demonstrate that both the dependency information and the prior knowledge are effective for CDR extraction, and the prior knowledge could significantly contribute to improve the performance.

The intra- and inter- sentence level CDR extraction are considered respectively in this paper, which ignores the unity of the document. In the future, we would like to explore it at document level to model the two level CDR in a unified framework.